\documentclass[letterpaper]{article}

\usepackage{aaai24} 
\usepackage{times} 
\usepackage{helvet} 
\usepackage{courier} 
\usepackage[hyphens]{url} 
\usepackage{graphicx}
\usepackage{natbib}  
\usepackage{caption} 
\usepackage{amsmath,amsfonts}
\usepackage{multirow}
\usepackage{booktabs}
\usepackage{array}
\newcommand{\STAB}[1]{\begin{tabular}{@{}c@{}}#1\end{tabular}}

\usepackage{xcolor}
\definecolor{url_color}{RGB}{237,2,140}

\title{DTL: Disentangled Transfer Learning for Visual Recognition}
\author {
    Minghao Fu,
    Ke Zhu,
    Jianxin Wu\thanks{J. Wu is the corresponding author.}
}
\affiliations {
    National Key Laboratory for Novel Software Technology, Nanjing University, China \\
School of Artificial Intelligence, Nanjing University, China\\
    fumh@lamda.nju.edu.cn, zhuk@lamda.nju.edu.cn, wujx2001@gmail.com
}

\begin{document}

\maketitle

\begin{abstract}
When pre-trained models become rapidly larger, the cost of fine-tuning on downstream tasks steadily increases, too. To economically fine-tune these models, parameter-efficient transfer learning (PETL) is proposed, which only tunes a tiny subset of trainable parameters to efficiently learn quality representations. However, current PETL methods are facing the dilemma that during training the GPU memory footprint is \emph{not} effectively reduced as trainable parameters. PETL will likely fail, too, if the full fine-tuning encounters the out-of-GPU-memory issue. This phenomenon happens because trainable parameters from these methods are generally entangled with the backbone, such that a lot of intermediate states have to be stored in GPU memory for gradient propagation. To alleviate this problem, we introduce Disentangled Transfer Learning (DTL), which disentangles the trainable parameters from the backbone using a lightweight Compact Side Network (CSN). By progressively extracting task-specific information with a few low-rank linear mappings and appropriately adding the information back to the backbone, CSN effectively realizes knowledge transfer in various downstream tasks. We conducted extensive experiments to validate the effectiveness of our method. The proposed method not only reduces a large amount of GPU memory usage and trainable parameters, but also outperforms existing PETL methods by a significant margin in accuracy, achieving new state-of-the-art on several standard benchmarks. The code is available at \textcolor{url_color}{https://github.com/heekhero/DTL}.

\end{abstract}

\section{Introduction}

The pipeline of large-scale pre-training plus fine-tuning has been popularized in various domains~\cite{bert,bart,mae,dino,mls,cab}. But traditional fine-tuning can be intractable due to GPU memory or time budget~\cite{mae}, since parameters of the entire large model have to be updated. Recently, parameter-efficient transfer learning (PETL) is proposed to update only a tiny subset of trainable parameters~\cite{adapter}. Because of its efficacy and the ability to prevent over-fitting, numerous variants~\cite{vpt,lora,noah,ssf,fact} of PETL successively emerged.

\begin{figure}
     \centering
         \includegraphics[width=0.45\textwidth]{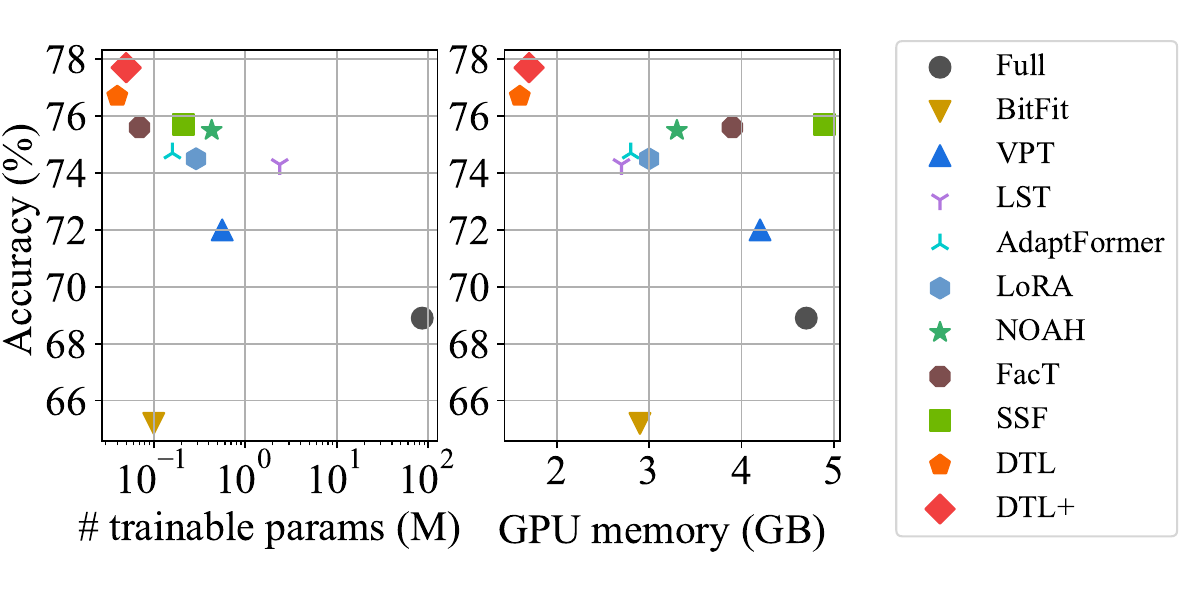}
         \caption{Top-1 accuracy on VTAB-1K~\cite{vtab} vs. different numbers of trainable parameters and GPU memory footprint. Our DTL achieves the highest accuracy with the least trainable parameters and GPU memory usage.}
         \label{fig:compare}
\end{figure}

Nevertheless, a huge decrease on trainable parameters does \emph{not} necessarily mean an equivalent reduction in GPU memory usage: the percentage of saved GPU memory is still small (around 25\%, cf. Fig.~\ref{fig:compare}). Even the PETL pipeline may still fail if a large model cannot be fine-tuned due to GPU memory shortage. This drawback is critical and fundamental. Hence, it is critical to devise a new method that effectively reduces GPU memory usage and fully explores the utility of large-scale pre-trained models.

\begin{figure*}
    \centering
    \includegraphics[width=0.75\linewidth]{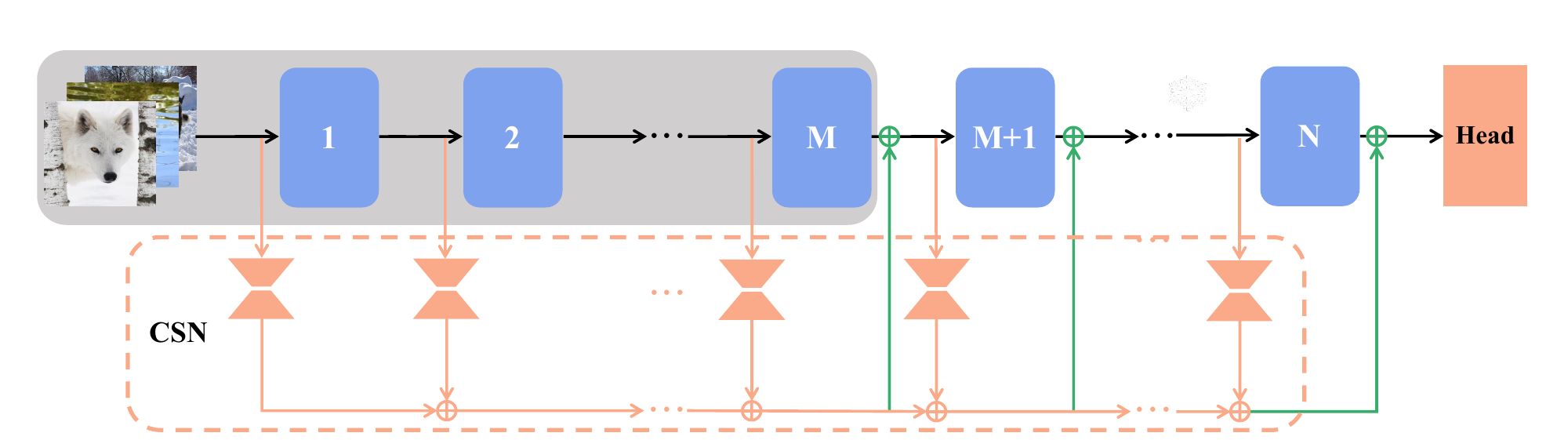}
    \caption{Illustration of DTL's network architecture for ViT~\cite{vit}. Our Compact Side Network (CSN) with scarce trainable parameters is plugged \emph{in parallel to} the backbone blocks. Specifically, before the forward calculation in each block, a low-rank linear mapping~\cite{lora} is applied to the input features to aggregate task-specific side information (orange arrows). This side information is added back to the output of later backbone blocks (green arrows) for adapting backbone features to downstream tasks. During fine-tuning, only parameters of the CSN module and the task-specific classification head are updated (illustrated in orange). Best viewed in color.}
    \label{fig:network}
\end{figure*}

One common characteristic of PETL methods~\cite{lora,adapter,vpt} is that \emph{they closely entangle the small trainable modules with the huge frozen backbone}. As indicated by~\citet{lst}, for a specific network parameter to be correctly updated, the model has to cache related intermediate gradients from activation values. This entangled design makes the cache a considerable part of GPU memory footprint, and thus hinders large pre-trained models from being applied in various tasks.

To address this fundamental drawback, we propose Disentangled Transfer Learning (DTL), which \emph{disentangles} the weights update from the backbone network by proposing a lightweight Compact Side Network (CSN). DTL not only greatly reduces GPU memory footage, but also achieves high accuracy in knowledge transfer (cf. Fig.~\ref{fig:compare}). 

As shown in Fig.~\ref{fig:network}, CSN composes of several low-rank linear mapping matrices to extract task-specific information, which is completely disentangled from the backbone. By injecting this information back to a few later backbone blocks, part of the intermediate features generated by pre-trained model are adaptively calibrated to make the features more discriminative for downstream tasks. We can also enhance DTL to DTL+, which inserts an additional global depthwise separable convolution (DWConv) layer~\cite{xception} to gather spatial information when injecting back from CSN to the backbone. DTL is very simple and compatible with various backbone architectures.

The output of early blocks in the backbone (covered by the gray region in Fig.\ref{fig:network}) is kept constant during fine-tuning, making it possible to reuse backbone features across \emph{multiple downstream tasks} when the same input is provided.

We conducted extensive experiments to verify the effectiveness of the proposed DTL, which achieved superior top-1 evaluation accuracy with significantly less trainable parameters and GPU memory during fine-tuning compared to its traditional PETL counterparts. Our contributions can be summarized as follows:
\begin{itemize}
    \item We analyze limitations of existing PETL methods from the perspective of GPU memory usage, which has a critical influence on the feasibility of fine-tuning.
    \item Motivated by our analysis, we propose DTL, a disentangled and simple framework for efficiently fine-tuning large-scale pre-trained models with significantly less trainable parameters and GPU memory usage.
    \item Extensive experiments are conducted to verify the effectiveness of DTL, which outperforms existing methods with a large margin.
\end{itemize}

\section{Related Work}

PETL adapts a large pre-trained model to downstream tasks in a parameter-efficient fashion. Now we present some typical PETL methods in both vision and language communities.

BitFit~\cite{bitfit} fine-tunes all bias terms in the backbone network to partially adapt pre-trained models to downstream tasks. VPT~\cite{vpt} introduces prompt-tuning, which prepends learnable tokens $P \in \mathbb{R}^{l \times d}$ to patch tokens $X \in \mathbb {R}^{n \times d}$ as $X' = [P, X]$ to act as the input of a ViT block. \citet{vpt} propose two variants: 1) VPT-Shallow, which only inserts $P$ before the first block; and 2) VPT-Deep, where the input of every block is concatenated with a different $P$. During fine-tuning, only $P$ along with the classification head $W$ are learnable.

Adapter~\cite{adapter} fine-tunes text Transformers~\cite{transformer,bert} with a bottleneck architecture consisting of a down projection layer $W_{down} \in \mathbb {R} ^{d \times d'}$ and an up projection layer $W_{up} \in \mathbb {R} ^{d' \times d}$. It's inserted after the Multi-Head Self-Attention (MHSA) and Feed-Forward Network (FFN). The computation is formulated as $X' = X + \Theta (X W_{down})W_{up}$, where $\Theta$ is the activation function and $X$ is the output of MHSA or FFN. By setting $d' \ll d$, the number of trainable parameters (\{$W_{down}$, $W_{up}$\}) is limited. AdaptFormer~\cite{adaptformer} further attaches the bottleneck to FFN in a parallel form:
\begin{equation}
    X' = X + \mathcal {FFN}(\mathcal {LN} (X)) + s \cdot  \Theta (X W_{down})W_{up} \,,
\end{equation}
where $X$ is the output of MHSA, $\mathcal {LN}$ is layer normalization~\cite{layernorm} and $s$ is a scalar factor.

SSF~\cite{ssf} linearly transforms the intermediate features $X$ of the backbone with scale $\gamma \in \mathbb {R}^d$ and shift $\beta \in \mathbb {R}^d$, as $X' = \gamma \odot X + \beta$, in which $\odot$ denotes element-wise product and $X$ comes from the output of all MHSA, FFN and LN operations. Like other approaches, backbone parameters are frozen while additional parameters \{$\gamma$,$\beta$\} are set to be learnable during fine-tuning.

LoRA~\cite{lora} decomposes the update of weights matrix $W$ in a linear layer with $W'=W + \Delta W$. $\Delta W \in \mathbb {R}^{d \times d}$ is implemented by a low-rank approximation using two matrices $A \in \mathbb {R}^{d \times r}$ and $B \in \mathbb {R}^{r \times d}$, with $\Delta W=A B$ and $r \ll d$. Then the output after fine-tuning becomes 
\begin{equation}
    X' = X W + X A B \,.
    \label{eq:lora}
\end{equation}
By integrating $A$ and $B$ into both query ($W_q$) and value ($W_v$) mapping matrices in MHSA respectively, LoRA achieves superior results over previous works.

FacT~\cite{fact} boosts the efficiency of low-rank tuning using tensorization-decomposition to store the update of trainable parameters, which contains two variants. The first one, termed as FacT-TT, decomposes $\Delta W$ as $\Delta W = s \cdot \Sigma \times_2 U^T \times_3 V^T$, where $U \in \mathbb{R}^{d \times r_1}$, $V \in \mathbb{R}^{d \times r_2}$, $\Sigma \in \mathbb{R}^{ 12 L \times r_1 \times r_2}$ and $\times_i$ is mode-$i$ product. The other one, FacT-TK, further pushes the decomposition as $\Delta W = s \cdot C \times_1 P^T \times_2 U^T \times_3 V^T$, where $U \in \mathbb{R}^{d \times r_2}$, $V \in \mathbb{R}^{d \times r_3}$, $P \in \mathbb{R}^{ 12 L \times r_1}$ and $C \in \mathbb {R}^{r_1 \times r_2 \times r_3}$, respectively. By setting $r_1,r_2,r_3 \ll d$, FacT is parameter-efficient.

NOAH~\cite{noah} tries to free up researchers from manual architecture design. They propose to firstly train a supernet including Adapter, VPT and LoRA modules. After that, an evolutionary algorithm is performed to search the reduction dimensionality $d'$ in Adapter, prompt length $l$ in VPT and rank $r$ in LoRA under the constraint on the number of trainable parameters.

\section{Limitations of Current PETL Methods}

Suppose there is an $N$ layer feed-forward network $y=f_N(f_{N-1}(...f_1(x)))$, where layer $i$ has a weight matrix $W_i$ and a bias term $b_i$. We denote $o_{i+1}$, $z_{i+1}$ as the output and pre-activation of layer $i$, respectively. Then, $o_{i+1}=\sigma(z_{i+1})=\sigma(W_i o_{i} + b_i)$, where $\sigma$ is the activation function. \citet{lst} shows that the gradients back propagated from the loss $L$ to $W_i$ and $b_i$ are
\begin{align}
        \frac{\partial L}{\partial W_i} &= \frac{\partial L}{\partial o_{i+1}} \sigma_i' o_i \,, \notag\\
        \frac{\partial L}{\partial b_i} &= \frac{\partial L}{\partial o_{i+1}} \sigma_i' \,,
		\label{eq:gradient_1}
\end{align}
where $\sigma_i'$ is the abbreviation of $\partial o_{i+1} / \partial z_{i+1}$. Furthermore, the term $\partial L / \partial o_{i+1}$ can be recursively expressed as
\begin{equation}
    \frac{\partial L}{\partial o_{i+1}} = \frac{\partial L}{\partial o_{i+2}} \frac{\partial o_{i+2}}{\partial z_{i+2}} \frac{\partial z_{i+2}}{\partial o_{i+1}} = \frac{\partial L}{\partial o_{i+2}} \sigma_{i+1}' W_{i+1} \,.
    \label{eq:gradient_2}
\end{equation}
To correctly calculate the gradients, except for parameters from the model (in this case, $W_i$ and $b_i$), \emph{all corresponding \{$\sigma_i'$\} in the chain rule have to be cached during fine-tuning}, which dominates the GPU memory usage.

We have introduced several representative PETL methods in the related work section, and \emph{all} these methods closely \emph{entangle} the trainable parameters with the backbone, which hardly reduces the GPU memory usage in caching \{$\sigma_i'$\}. This property shows that the GPU memory footprint cannot be effectively reduced compared to a full fine-tuning, even though the number of trainable parameters is very small.

To solve this fundamental difficulty, we propose a new learning paradigm called Disentangled Transfer Learning (DTL). The central idea of DTL is to \emph{disentangle} the weight updating of the small extra modules from the the backbone network (cf. Fig.~\ref{fig:network}). Therefore, the relevant $\sigma_i'$ stored for back propagation can be drastically reduced (cf. Eq.~\ref{eq:gradient_1} and~\ref{eq:gradient_2}). In this way, DTL successfully pushes the limits of current PETL further from not only being parameter-efficient but also reducing the necessary GPU memory size in fine-tuning large-scale pre-trained models.

\section{Method}

We propose a \emph{disentangled}, \emph{simple} and \emph{effective} approach to fine-tune large-scale pre-trained models properly. In order to trade off the recognition accuracy and architectural complexity in different environments, we introduce two variants of our method, termed as DTL and DTL+. 

\subsection{DTL: Simplicity Matters}

We first show the simplest version of our solution. In Fig.~\ref{fig:network} we illustrate the pipeline of the proposed architecture for the ViT~\cite{vit} backbone, which is mainly built up with a \emph{Compact Side Network} (CSN). CSN is plugged into the backbone for information aggregation and feature adaptation. Note that the proposed method is compatible with other types of backbones, which will be discussed soon.

Given a ViT backbone containing $N$ blocks, the forward calculation can be formulated as $z=b_N(b_{N-1}( ... b_1(x)))$, where $b_i$ is the $i$-th block, $x\in \mathbb R^{(n+1)\times d}$ is the input tokens (patch tokens plus one \emph{cls} token) and $z\in \mathbb R^{(n+1)\times d}$ is the output tokens, respectively. Denote $z_{i+1}$ as the output of $b_i$, hence $z_{i+1}=b_i(z_i)$ and $z_1=x$. Our CSN composes of $N$ low-rank linear transformation matrices~\cite{lora}, with each being plugged into one block to extract task-specific information. Denote $w_i=a_i c_i \in \mathbb R^{d \times d}$ as the weight matrix accounting for the $i$-th block, with $a_i \in \mathbb R^{d \times d'}$, $c_i \in \mathbb R^{d' \times d}$ and $d' \ll d$, CSN progressively gathers information from each block as
\begin{align}
    h_{i+1} &= h_i + z_i w_i \,, \\
    z_{i+1} &= b_i (z_i) \,,
\end{align}
where $h_{i+1}$ is the output of the $i$-th layer of CSN ($h_1=0$). After that, starting from the $M$-th block, the aggregated task-specific information $h_{i+1}$ is used to adapt $z_{i+1}$ to downstream tasks by adding it back to $z_{i+1}$. Hence, when $i \geq M$,
\begin{equation}
        z_{i+1}' = z_{i+1} + \theta(h_{i+1}) \,,
\end{equation}
where $z_{i+1}'$ is the adapted output of $b_i$ and $\theta$ is the Swish activation~\cite{swish}, where $\theta(x)=\frac{x}{1+e^{-\beta x}}$. To prevent $z_{i+1}'$ from drastically shifting away from $z_{i+1}$ at the beginning of fine-tuning, $a_i$ is initialized following a uniform distribution and $c_i$ is zero-initialized. To sum up, the output from the $i$-th block is
\begin{equation}
        z_{i+1}' = \left\{\begin{aligned}
            &z_{i+1} + \theta(h_{i+1}) &&\text{if} \ \ i \geq M \\
            &z_{i+1} &&\text{otherwise} \,.
        \end{aligned}\right.
		\label{eq:dtl_full}
\end{equation}

We find that a small $d'$ (2 or 4) performs fairly well, which suggests high redundancy in the backbone features. Therefore, in addition to keep $d'$ small, we use a large $\beta$ (100) in Swish (i.e., $\theta$) to further reduce the redundancy. Consequently, about half of $\theta(h_{i+1})$ is close to zero.

\subsection{DTL+ : Effectiveness Matters}

To further boost the effectiveness of the proposed method, we append an additional global depthwise separable convolution (DWConv) layer~\cite{xception} $g$ to each side layer after $\theta$ is applied. The formulation of DTL+ is
\begin{equation}
    z_{i+1}' = \left\{\begin{aligned}
		&z_{i+1} + g(\theta(h_{i+1})) &&\text{if} \ \ i \geq M  \\
		&z_{i+1} &&\text{otherwise} \,.
	\end{aligned}\right.
	\label{eq:dtl+_full}
\end{equation}
The stride of $g$ is set to 1 and zero-padding is used to ensure that $g$ does not change feature size. Note that \emph{$g$ is shared across different CSN layers}, so that the number of trainable parameters in $g$ is small compared to the initial CSN, and the whole CSN module is still lightweight. The introduction of $g$ makes spatial information properly processed by our CSN module. With this operation, it's easier for the model to recognize new categories.

\subsection{Advantages}

The proposed approach has some significant advantages, which we discuss explicitly.

\textbf{Disentangled.}
As shown in Fig.~\ref{fig:network}, the proposed CSN is a plugin mostly detached from the backbone, which interacts with the backbone in a plug-and-play manner. This characteristic makes our method easy to implement, and is \emph{compatible with almost all backbone networks}. Modern deep neural networks are mostly divided into several intermediate stages and the feature dimensionality within one stage is the same. By re-initializing the hidden state of CSN $h_i$ to 0 at the beginning of each stage, our method can be easily transferred to different backbone architectures.

From the perspective of GPU memory usage, in previous methods weight update is directly entangled with the backbone. As we have analyzed before, although the number of trainable parameters is small, they still require a lot of GPU memory to cache many \{$\sigma_i'$\} for gradient propagation. Our method alleviates this issue by 1) separating the forward pass of the backbone from CSN; and 2) only entangling them at late stages ($i \geq M$). Within our framework, \emph{no gradients are back propagated to the first $M$ blocks in the backbone} (the gray region in Fig.~\ref{fig:network}). Hence, the number of cached \{$\sigma_i'$\} is drastically reduced in CSN, resulting in a highly efficient way to realize GPU memory reduction.

Finally, we discuss another advantage drawn from our disentangled architecture: the possibility of feature reuse. Consider a scenario where we need to perform different tasks on one input image (e.g., simultaneously predict age and gender for a human). We have several fine-tuned models, and in previous methods the intermediate features $z_{i+1}$ generated by these fine-tuned models are \emph{different to each other}. In other words, there is no way to share computation in the backbone across different tasks. Therefore a standard process is learning a group of task-specific parameters for each task (cf. Table~\ref{tab:unit}) and conducting each task individually.

\begin{table}
	\centering
	\small
	\begin{tabular}{lcr}
	\toprule[1.5pt]
	Method& Source & \#unit \\
	\hline
	\specialrule{0em}{1pt}{1pt}
	LoRA & low-rank matrices in $W_q$, $W_v$ & 24 \\
	NOAH & low-rank matrices, bottlenecks, prompts & 36\\
	FacT & decomposed tensors & 144 \\
	SSF & pairs of $\gamma$, $\beta$ & 148 \\
	\hline
	DTL & low-rank matrices & 12 \\
	DTL+ & low-rank matrices, DWConv & 13 \\
	\bottomrule[1.5pt]
	\end{tabular}
	\caption{Statistics of number of minimal structural units in different methods. ``Source'' denotes the types of minimal structural units. ``\#unit'' denotes the number of minimal structural units in the backbone.}
	\label{tab:unit}
\end{table}

Conversely, as shown in the gray region of Fig.~\ref{fig:network}, in our DTL the intermediate features before block $M$ remain the same after fine-tuning, such that we can share part of the backbone computation between different tasks. 

We take the 19 datasets in VTAB-1K~\cite{vtab} to illustrate the above-described situation. We fine-tune to obtain 19 models, and assume that we need to get all 19 classification results for the one input image using all these 19 models. The goal is to check how much speedup can be achieved during inference. Firstly, we feed the same input image into 19 different models after fine-tuning with LoRA~\cite{lora}, which acts as the baseline. Then we implement our method to simultaneously conduct 19 tasks but with the backbone feature shared in the first 6 blocks (by setting $M=7$). Experimental results show that approximately \emph{45\% inference latency is saved during inference}.

\textbf{Simple}. Since the CSN is disentangled from the backbone network, our method naturally shows higher simplicity compared to previous methods. Since all PETL methods add various types of structural units as extra trainable parameters, to verify the simplicity of our DTL in more detail, we compare the number of such minimal structural units of our method with previous methods in Table~\ref{tab:unit}.

In this context, the phrase \emph{minimal structural unit} means the atomic modules inserted into the backbone network. For example, in LoRA~\cite{lora}, one minimal structural unit comprises of a pair of $A$ and $B$ matrices to constitute $\Delta W$ (cf. Eq.~\ref{eq:lora}). Since it inserts $\Delta W$ into both $W_q$ and $W_v$ for MHSA in every Transformer block, the total number of these units is 24. It is similarly defined for other methods as well, which include: 1) pairs of $\gamma$ and $\beta$ in SSF; 2) the modules to be searched in supernet and maintained in subnet in NOAH; 3) decomposed tensors in FacT; 4) pairs of matrices $a_i$ and $c_i$ in our DTL; 5) the additional global DWConv layer in DTL+. As shown in Table~\ref{tab:unit}, the proposed method requires much fewer minimal structural units compared to existing methods. 

We notice that a previous work LST~\cite{lst} also has a side network design. However, their architecture is very complicated and requires sophisticated techniques~\cite{structpruning} to initialize, resulting in the existence of large number of trainable parameters as shown in Table~\ref{tab:vtab} (about 50$\times$ compared to ours). As analyzed before, we reduce the fine-tuning redundancy by setting $d'$ to a very small value (2 or 4), which is far less than previous counterparts (e.g., 8 in LoRA and Adapter). This choice makes our DTL not only simple in structure, but also contains much fewer trainable parameters than other methods.

\textbf{Effective.} We conducted extensive experiments to verify the effectiveness of the proposed method. The results demonstrate that our method shows superior recognition accuracy across multiple architectures, achieving new state-of-the-art on several standard benchmarks.

\begin{table*}
	\centering
	\setlength{\tabcolsep}{0.3pt}
	\small
	\begin{tabular}{p{2.2cm}<{}p{0.65cm}<{\centering}p{0.65cm}<{\centering}|p{0.65cm}<{\centering}p{0.65cm}<{\centering}p{0.65cm}<{\centering}p{0.65cm}<{\centering}p{0.65cm}<{\centering}p{0.65cm}<{\centering}p{0.65cm}<{\centering}|p{0.65cm}<{\centering}p{0.65cm}<{\centering}p{0.65cm}<{\centering}p{0.65cm}<{\centering}|p{0.65cm}<{\centering}p{0.65cm}<{\centering}p{0.65cm}<{\centering}p{0.65cm}<{\centering}p{0.65cm}<{\centering}p{0.65cm}<{\centering}p{0.65cm}<{\centering}p{0.65cm}<{\centering}|p{0.65cm}<{\centering}}
	\toprule[1.5pt]
	\multicolumn{3}{c|}{}&\multicolumn{7}{c|}{\textbf{Natural}}&\multicolumn{4}{c|}{\textbf{Specialized}}&\multicolumn{8}{c|}{\textbf{Structured}}&\\
	&\multicolumn{1}{c}{\STAB{\rotatebox[origin=c]{90}{\#param (M)}}}
	&\multicolumn{1}{c|}{\STAB{\rotatebox[origin=c]{90}{GPU mem (GB)}}}
	&\multicolumn{1}{c}{\STAB{\rotatebox[origin=c]{90}{Cifar100}}}
	&\multicolumn{1}{c}{\STAB{\rotatebox[origin=c]{90}{Caltech101}}}
	&\multicolumn{1}{c}{\STAB{\rotatebox[origin=c]{90}{DTD}}}
	&\multicolumn{1}{c}{\STAB{\rotatebox[origin=c]{90}{Flower102}}}
	&\multicolumn{1}{c}{\STAB{\rotatebox[origin=c]{90}{Pets}}}
	&\multicolumn{1}{c}{\STAB{\rotatebox[origin=c]{90}{SVHN}}}
	&\multicolumn{1}{c|}{\STAB{\rotatebox[origin=c]{90}{Sun397}}}
	&\multicolumn{1}{c}{\STAB{\rotatebox[origin=c]{90}{Camelyon}}}
	&\multicolumn{1}{c}{\STAB{\rotatebox[origin=c]{90}{EuroSAT}}}
	&\multicolumn{1}{c}{\STAB{\rotatebox[origin=c]{90}{Resisc45}}}
	&\multicolumn{1}{c|}{\STAB{\rotatebox[origin=c]{90}{Retinopathy}}}
	&\multicolumn{1}{c}{\STAB{\rotatebox[origin=c]{90}{Clevr-Count}}}
	&\multicolumn{1}{c}{\STAB{\rotatebox[origin=c]{90}{Clevr-Dist}}}
	&\multicolumn{1}{c}{\STAB{\rotatebox[origin=c]{90}{DMLab}}}
	&\multicolumn{1}{c}{\STAB{\rotatebox[origin=c]{90}{KITTI-Dist}}}
	&\multicolumn{1}{c}{\STAB{\rotatebox[origin=c]{90}{dSpr-Loc}}}
	&\multicolumn{1}{c}{\STAB{\rotatebox[origin=c]{90}{dSpr-Ori}}}
	&\multicolumn{1}{c}{\STAB{\rotatebox[origin=c]{90}{sNORB-Azim}}}
	&\multicolumn{1}{c|}{\STAB{\rotatebox[origin=c]{90}{sNORB-Ele}}}
	&\multicolumn{1}{c}{\STAB{\rotatebox[origin=c]{90}{Average}}}\\
	\specialrule{0em}{1pt}{1pt}
	\hline
	\specialrule{0em}{1pt}{1pt}
	\multicolumn{22}{l}{\emph{Traditional Fine-Tuning}}\\
	\hline
	\specialrule{0em}{1pt}{1pt}
	Full&85.8&4.7 &68.9&87.7&64.3&97.2&86.9&87.4&38.8&79.7&95.7&84.2&73.9&56.3&58.6&41.7&65.5&57.5&46.7&25.7&29.1&68.9 \\
	Linear&0&0.6&64.4&85.0&63.2&97.0&86.3&36.6&51.0&78.5&87.5&68.5&74.0&34.3&30.6&33.2&55.4&12.5&20.0&9.6&19.2&57.6\\
	\hline
	\specialrule{0em}{1pt}{1pt}
	\multicolumn{22}{l}{\emph{PETL methods}}\\
	\hline
	\specialrule{0em}{1pt}{1pt}
	BitFit&0.10&2.9&72.8&87.0&59.2&97.5&85.3&59.9&51.4&78.7&91.6&72.9&69.8&61.5&55.6&32.4&55.9&66.6&40.0&15.7&25.1&65.2\\
	VPT&0.56&4.2&78.8&90.8&65.8&98.0&88.3&78.1&49.6&81.8&96.1&83.4&68.4&68.5&60.0&46.5&72.8&73.6&47.9&32.9&37.8&72.0 \\
	LST&2.38&2.7&59.5&91.5&69.0&99.2&89.9&79.5&54.6&86.9&95.9&85.3&74.1&81.8&61.8&52.2&81.0&71.7&49.5&33.7&45.2&74.3\\
	LoRA&0.29&3.0&67.1&91.4&69.4&98.8&90.4&85.3&54.0&84.9&95.3&84.4&73.6&\bf82.9&\bf69.2&49.8&78.5&75.7&47.1&31.0&44.0&74.5\\
	AdaptFormer&0.16&2.8&70.8&91.2&70.5&99.1&90.9&86.6&54.8&83.0&95.8&84.4&\bf76.3&81.9&64.3&49.3&80.3&76.3&45.7&31.7&41.1&74.7 \\
	NOAH&0.43&3.3&69.6&92.7&70.2&99.1&90.4&86.1&53.7&84.4&95.4&83.9&75.8&82.8&68.9&49.9&81.7&81.8&48.3&32.8&44.2&75.5\\
	FacT&0.07&3.9&70.6&90.6&70.8&99.1&90.7&88.6&54.1&84.8&96.2&84.5&75.7&82.6&68.2&49.8&80.7&80.8&47.4&33.2&43.0&75.6\\
	SSF&0.21&4.9&69.0&92.6&\bf75.1&99.4&91.8&90.2&52.9&87.4&95.9&87.4&75.5&75.9&62.3&\bf53.3&80.6&77.3&54.9&29.5&37.9&75.7  \\
	\hline
	\specialrule{0em}{1pt}{1pt}
	DTL&\bf0.04&\bf1.6&69.6&94.8&71.3&99.3&91.3&83.3&56.2&87.1&96.2&86.1&75.0&82.8&64.2&48.8&81.9&93.9&53.9&34.2&47.1&76.7\\
	DTL+&0.05&1.7&70.4&\bf95.1&71.5&\bf99.4&\bf91.8&87.5&56.8&87.7&96.6&86.9&74.7&81.6&65.1&51.3&\bf82.3&\bf97.2&\bf54.9&\bf36.0&49.3&77.7 \\
	DTL+*&0.05&3.1&\bf74.1&94.8&71.8&99.4&91.7&\bf90.4&\bf57.2&\bf87.9&\bf96.7&\bf87.5&74.8&81.9&64.7&51.5&81.9&93.9&54.0&35.6&\bf50.3&\bf77.9\\
	\bottomrule[1.5pt]
	\end{tabular}
	\caption{Results on the VTAB-1K benchmark with ViT-B/16 as the backbone. ``\#param'' denotes the number of trainable parameters. ``GPU mem'' specifies the peak GPU memory footprint when fine-tuning with batch size 32. ''Average'' is the group-wise average accuracy over three groups. The best results among PETL methods are in bold face.}
	\label{tab:vtab}
\end{table*}

\section{Experiments}

We conducted thorough experiments to evaluate the proposed method. First, we present results on the VTAB-1K~\cite{vtab} benchmark with two prevalent backbones, ViT-B/16~\cite{vit} and Swin-B~\cite{swin}. Then we verify the generalization ability of our method on few-shot learning and domain generalization. Finally, ablation studies are conducted for further analysis.

\subsection{Implementation Details} 

Following previous work~\cite{ssf,fact}, we take AdamW~\cite{adamw} with cosine learning rate schedule as the optimizer. $\beta$ in Swish is fixed to 100. All pre-trained models are fine-tuned by 100 epochs with batch size 32. The rank $d'$ of low-rank linear mappings in CSN is 2 for ViT and 4 for Swin-B. We set $M$ (cf. Eq.~\ref{eq:dtl_full}-\ref{eq:dtl+_full}) of DTL and DTL+ as 7 for the ViT backbone, which means half of the later blocks' output is calibrated by adding back the output from CSN. It is similarly defined for Swin-B with half of the layers adapted as well. Note that unlike previous methods~\cite{noah,ssf}, except for the standard data augmentation, we do \emph{not} use any additional tricks such as mixup~\cite{mixup}, cutmix~\cite{cutmix} or label smoothing~\cite{label_smooth}. More details are available at \url{https://www.lamda.nju.edu.cn/fumh/files/DTL/DTL_appendix.pdf}.

\subsection{Experiments on VTAB-1K}

\textbf{Datasets.} VTAB-1K was introduced by~\citet{vtab} to evaluate the generalization ability of representation learning approaches. It contains diverse images from 19 different datasets, grouped as 1) \emph{Natural} images captured by standard cameras; 2) \emph{Specialized} images captured by specialist equipment; and 3) \emph{Structured} images generated in simulated environments. They vary in task-specific objectives (e.g., classic visual recognition, object counting or depth prediction) and there are only 1,000 images in each dataset for training. It is a challenging benchmark to evaluate PETL methods. We report the top-1 recognition accuracy on the test set.

\textbf{Baseline methods.}  First, two traditional fine-tuning techniques are included in all experiments. One is `Full', which fine-tunes the entire pre-trained model. The other is `Linear', which only fine-tunes task-specific classification head. Second, we choose BitFit~\cite{bitfit}, VPT~\cite{vpt}, LST~\cite{lst}, AdaptFormer~\cite{adaptformer}, LoRA~\cite{lora}, NOHA~\cite{noah}, FacT~\cite{fact} and SSF~\cite{ssf} as PETL baselines. We follow the setting in~\cite{noah,ssf} to report the results for a fair comparison.

In addition to DTL and DTL+ ($M=7$), we further extend DTL+ where all of the blocks are adapted (i.e., $M=1$, denoted as `DTL+*').

\textbf{Main results.} Results on ViT-B/16 are shown in Table~\ref{tab:vtab}. Our DTL shows a 1.0\% gain on average accuracy compared to the previous state-of-the-art method SSF. By integrating a global DWConv, DTL+ further increases the average improvement to 2.0\%. Specifically, DTL+ reaches the best top-1 accuracy on 11 out of 19 datasets, where the improvements compared to SSF range from 0.3\% to 19.9\%. Even if the dataset `dSpr-Loc' with the most significant gain of 19.9\% is removed, DTL+ is still far ahead on average accuracy of the remaining 18 datasets and outperforms SSF by 1.3\%.

DTL only introduces 0.04M trainable parameters, which is 43\% less compared to FacT. DTL+ specifies a few more trainable parameters (+0.01M) because of the introduction of a shared DWConv, but is still significantly less than previous PETL methods. Moreover, DTL and DTL+ consume 1.6GB and 1.7GB GPU memory during fine-tuning, respectively, which is far less than other PETL baselines. Compared to full fine-tuning, the GPU memory saving rate is about 65\% on average (or saving roughly two thirds). 

Another interesting observation is DTL+ and DTL+* show different distributions on accuracy improvements between three groups. For the `Natural' group with smaller domain discrepancy between pre-trained models and downstream tasks, DTL+* outperforms DTL+ by 1\%. For the `Structured' group with a large domain gap, DTL+ surpasses DTL+* by 0.5\% instead. This observation also appears similarly in Table~\ref{tab:swin} on Swin-B. We thus conjecture that since DTL+* has larger capacity than DTL+, it is prone to over-fitting when facing large domain gaps.

We observe that DTL+* shows more improvements in top-1 accuracy compared to DTL+, which indicates that the feature adaptation in early blocks within backbone is, in general, useful for transfer learning. However, the GPU memory usage in this case is increased to 3.1 GB, yet with only 0.2\% accuracy gain compared to DTL+, implying the limited cost-efficiency of adapting early features.

As presented in Table~\ref{tab:swin}, the performance of DTL and DTL+ on Swin-B shows similar trends with ViT. DTL+ achieves new state-of-the-art, outperforming FacT with a significant margin of 1\% on average accuracy. DTL keeps the least trainable parameters and GPU memory footprint. Compared to full fine-tuning, DTL drastically saves GPU memory usage by 75\%.

\begin{table}
	\centering
	\small
	\setlength{\tabcolsep}{1pt}
	\begin{tabular}{p{1.4cm}<{}p{0.7cm}<{\centering}p{0.7cm}<{\centering}p{0.7cm}<{\centering}p{0.7cm}<{\centering}p{0.7cm}<{\centering}p{0.7cm}<{\centering}}
	\toprule[1.5pt]
	Method &\#p & \#m &Nat.&Spe.&Str.&Avg.\\\hline
	\specialrule{0em}{1pt}{1pt}
	Full&86.7&6.1 &79.2&86.2&59.7&75.0\\
	Linear&0&0.9 &73.5&80.8&33.5&62.6\\
	BitFit&0.20 &3.7&74.2&80.1&42.4&65.6\\
	VPT& 0.16 &4.6&76.8&84.5&53.4&71.6\\
	FacT&0.14 &5.6&83.1&86.9&62.1&77.4\\
	DTL &\bf0.09&\bf1.5&82.4&\bf87.0&64.2&77.9\\
	DTL+ & 0.13 & 1.6 & 82.4 & 86.8 & \bf66.0 &  78.4 \\
	DTL+* & 0.14 & 4.0& \bf83.2 & 87.0 & 65.7 & \bf78.6 \\
	\bottomrule[1.5pt]
	\end{tabular}
	\caption{Results on VTAB-1K with Swin-B backbone. `\#p' is the number of trainable parameters. `\#m' is peak GPU memory footprint in fine-tuning. Nat./Spe./Str./Avg. are the results in three VTAB groups and their group-wise average.}
	\label{tab:swin}
\end{table}

\begin{figure*}
	\centering
		\includegraphics[width=0.995\textwidth]{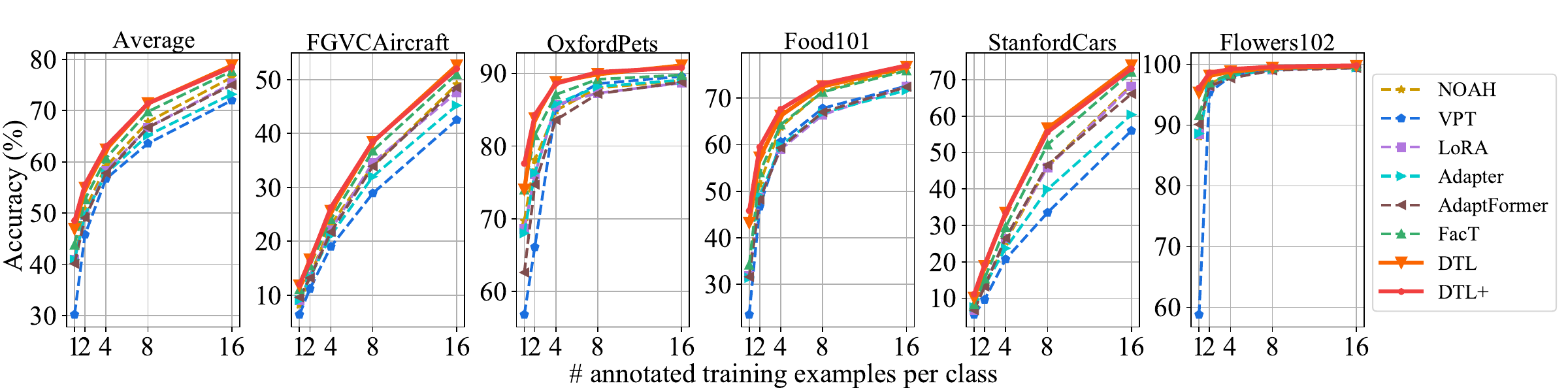}
		\caption{Top-1 accuracy on fine-grained few-shot benchmark with ViT-B/16 as the backbone. Best viewed in color. Note that our approach with less trainable parameters and GPU memory footprint outperforms all baseline methods.}
		\label{fig:fsl}
\end{figure*}

\subsection{Experiments on Few-shot Learning}

\textbf{Datasets.} Now we further evaluate on five fine-grained few-shot learning benchmark: Aircraft~\cite{fgvc_aircraft}, Pets~\cite{pets}, Food-101~\cite{food_101}, Cars~\cite{stanford_cars} and Flowers102~\cite{flowers_102}. Following~\citet{fact}, we fine-tune the pre-trained model with training set containing \{1, 2, 4, 8, 16\}-shot per class and report the average accuracy on test set over 3 seeds. 

\textbf{Main results.} As illustrated in Fig.~\ref{fig:fsl}, the proposed DTL and DTL+ outperform all baseline PETL methods in all cases. Furthermore, we observe that the average improvements of DTL+ compared to previous state-of-the-art FacT across different shots are gradually decreased from 4.7\% in 1-shot to 0.8\% in 16-shot, which reveals that our method is consistently effective, especially in low-data regimes.

\subsection{Experiments on Domain Generalization}

\textbf{Datasets.} We follow~\citet{noah} to conduct experiments on domain generalization to evaluate the robustness of our method when domain shift~\cite{domain_generalization_survey} is inevitable. In this scenario, the training set to fine-tune the pre-trained ViT-B/16 model is sampled from the original training set of ImageNet-1K, with each class containing 16 shot of images. Apart from the validation set of ImageNet-1K, the model is evaluated on 4 datasets, which are 1) ImageNet-Sketch~\cite{imagenet_sketch} composed of sketch images sharing the same label space with ImageNet-1K, 2) ImageNet-V2~\cite{imagenet_v2} collected from different sources compared with ImageNet-1K, 3) ImageNet-A~\cite{imagenet_a} consisting of adversarial examples, and 4) ImageNet-R~\cite{imagenet_r} containing various artistic renditions of ImageNet-1K. The reported accuracy is average by 3 different random seeds.

\begin{table}
	\centering
	\small
    \begin{tabular}{lccccc}
    \toprule[1.5pt]
    \multirow{2}{*}{Method} & \textbf{Source} & \multicolumn{4}{c}{\textbf{Target}} \\ 
    \cmidrule(lr){2-2} \cmidrule(lr){3-6}
    & ImageNet & -Sketch & -V2 & -A & -R \\
    \midrule
    Adapter &70.5 & 16.4 & 59.1 & 5.5 & 22.1 \\
    VPT & 70.5 & 18.3 & 58.0 &  4.6 & 23.2 \\
    LoRA & 70.8 & 20.0 & 59.3 &  6.9 & 23.3 \\
    NOAH & 71.5 & 24.8 & 66.1 &  11.9 & 28.5  \\
    \midrule
    DTL &78.3 & 35.4& 67.8& 14.0& 34.4\\
    DTL+ & \bf78.7 & \bf35.7 & \bf67.8 & \bf14.2 & \bf34.4 \\
    \bottomrule[1.5pt]
    \end{tabular}
    \caption{Top-1 accuracy on domain generalization experiments with ViT-B/16 as the backbone. Our method shows significant gains w.r.t baseline methods.}
    \label{tab:dg}
\end{table}

\textbf{Main results.} The results of domain generalization experiments are shown in Table~\ref{tab:dg}. We observe that compared to previous state-of-the-art method NOAH, DTL and DTL+ achieve impressive gains in evaluation accuracy, especially on ImageNet, ImageNet-Sketch and ImageNet-R, where the average improvement is up to about 8\%. These comparisons show excellent robustness of DTL and DTL+ for alleviating the domain shift problem and well demonstrate the effectiveness of the proposed method together with previous experiments.

\subsection{Ablation Studies}

\begin{figure}
	\centering
		\includegraphics[width=0.3\textwidth]{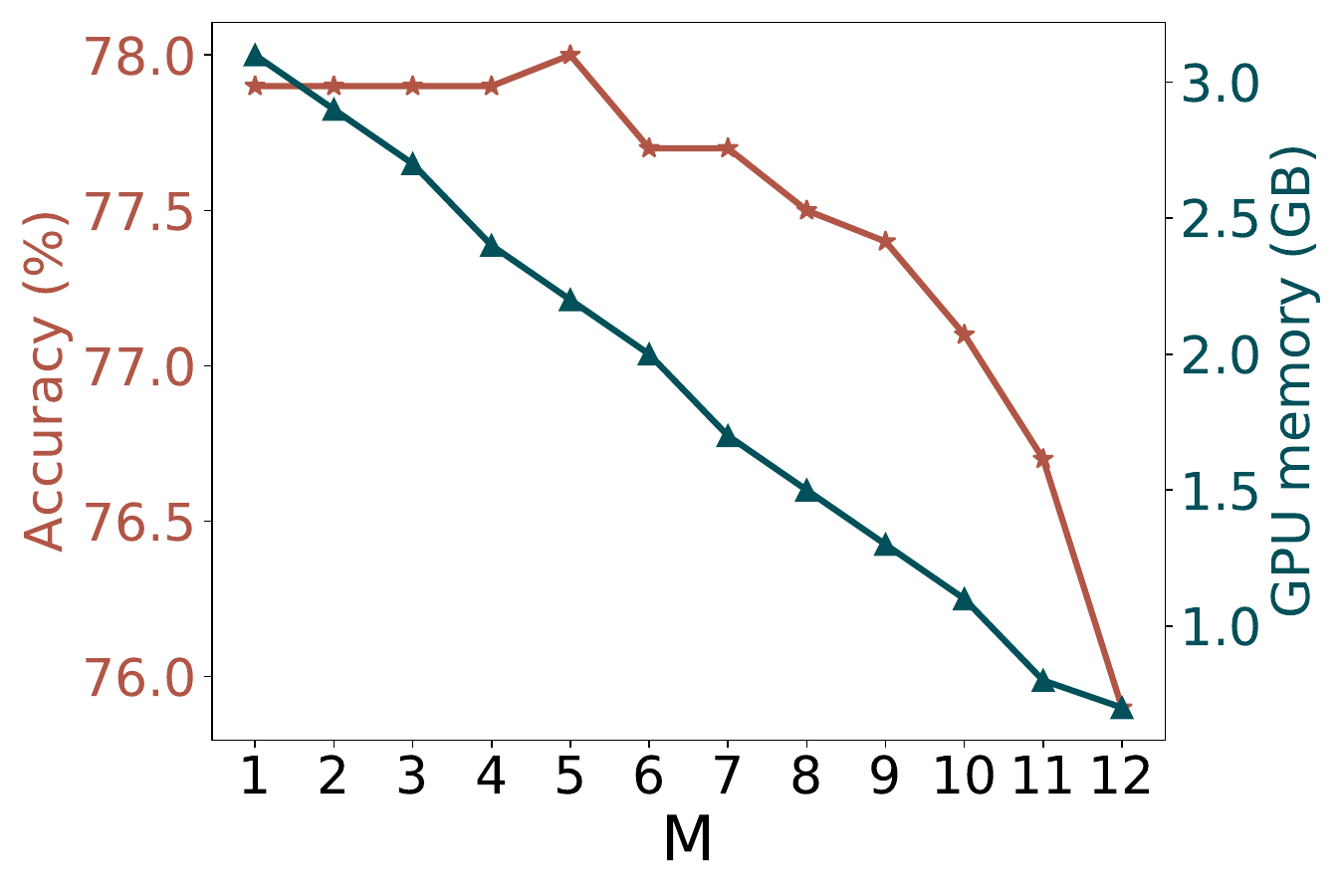}
		\caption{Top-1 accuracy and peak GPU memory footprint under various $M$ in Eq.~\ref{eq:dtl+_full}. Our method is consistently effective across different $M$.}
		\label{fig:aab}
\end{figure}

\textbf{Sensitivity to $M$.} In DTL and DTL+, the beginning index ($M$) of blocks in the backbone to add back the output of CSN for feature adaptation is critical for final performance. In Fig.~\ref{fig:aab} we plot the curve of accuracy and GPU memory footprint by varying $M$. There is a clear trend that the number of GPU memory usage decreases almost linearly as $M$ becomes larger. By decreasing $M$ from 12 to 11, the recognition accuracy is significantly boosted to 76.7\% from 75.9\%, and the improvements gradually saturate when $M < 6$, implying the feasibility and effectiveness of feature sharing as we described in the methods section. Finally, the range of accuracy across different $M$ is 2.1\%, indicating that the recognition accuracy is not sensitive to the exact value of $M$, so by default we set $M=7$ to pursue the best trade-off between effectiveness and efficiency.

\begin{table}
\centering
\small
    \begin{tabular}{lcccc}
    \toprule[1.5pt]
    $d'$ & Swish ($\theta$) & DWConv ($g$) & Avg. \\
    \midrule
    2 & & & 76.0 \\
    2 & \checkmark & & 76.7 \\
    2 & \checkmark & \checkmark & 77.7 \\
    4 & \checkmark & \checkmark & 77.6 \\
    1 & \checkmark & \checkmark & 77.2 \\
    \bottomrule[1.5pt]
    \end{tabular}
    \caption{Ablation results by varying different architectural choices, where the second and third lines denote default DTL and DTL+, respectively.}
    \label{tab:module}
\end{table}
 
\textbf{Modular ablation.} In Table~\ref{tab:module}, we provide ablation results on the VTAB-1K benchmark from ViT-B/16. The first line, where the layer-wise output of CSN with rank ($d'=2$) is directly added back to backbone (i.e, $z_{i+1}' = z_{i+1} + h_{i+1}$ when $i \geq M$), is the baseline. It achieves a 76.0\% average accuracy. By progressively integrating the Swish activation function $\theta$ in DTL and a global DWConv $g$ in DTL+, the accuracy is consistently improved. For DTL+, a higher ($d'=4$) or lower ($d'=1$) rank both make the accuracy worse than the default ($d'=2$). Interestingly even when the trainable parameters of CSN are extremely limited with $d'=1$, our DTL+ is still more effective than previous PETL methods.

\begin{table}
\centering
\small
    \begin{tabular}{lrcc}
    \toprule[1.5pt]
    \multirow{2}{*}{Methods} & \multicolumn{3}{c}{Throughput (imgs/sec)} \\ 
    \cline{2-4}
    \specialrule{0em}{1pt}{1pt}
    & bs=1       & bs=4  & bs=16      \\ 
    \midrule[0.4pt]
    Full & 161 & 636 & 952 \\
    \midrule[0.4pt]
    LST & 71 &	279 &  729 \\
    NOAH & 79 &	306 &  798 \\
    AdaptFormer & 108 &	436 & 876 \\
    DTL+ & 120 & 469 & 877 \\
    DTL & 131 & 528 & 892 \\
    \bottomrule[1.5pt]
    \end{tabular}
    \caption{Throughput (number of images processed per second with ViT-B/16 as the backbone) measured on a single NVIDIA 3090 GPU with mixed precision inference.}
\label{tab:efficiency}
\end{table}

\textbf{Inference efficiency.}
We further study the efficiency of our method during inference by comparing the throughput with some baselines. As illustrated in Table~\ref{tab:efficiency}, thanks to its simplicity, the empirical throughput of DTL+ is consistently higher than previous PETL counterparts. The simplest version of our method, DTL, boosts the inference efficiency a step further and significantly shows more speedup compared to traditional PETL methods.

\section{Conclusions and Limitations}

In this paper, we proposed Disentangled Transfer Learning, a new paradigm for fine-tuning large-scale pre-trained models. To trade off the efficiency and effectiveness, we designed two variants, DTL and DTL+. The most important property of DTL is, by disentangling weights update of trainable parameters from the backbone, it drastically reduces the GPU memory footprint required during fine-tuning. At the same time, the proposed method contains less trainable parameters and achieves competitive or even better accuracy compared to traditional PETL methods. Extensive experiments on several standard benchmarks plus ablations clearly show that our method is not only effective but also efficient for fine-tuning, indicating its great potential for practical usage.

An obvious limitation of DTL is that its granularity of interaction between the backbone and the trainable modules is very coarse. This is caused by the disentangled design. In the future, a better trade-off may be made between the two desired properties: disentanglement and interaction.

\section{Acknowledgments}
This research was partly supported by the National Natural Science Foundation of China under Grant 62276123 and Grant 61921006.

\bibliography{aaai24}

\end{document}